\pdfoutput=1

\documentclass[11pt]{article}

\usepackage{acl}

\usepackage{times}
\usepackage{latexsym}
\usepackage{graphicx}
\usepackage{amsmath}
\usepackage{amssymb}
\usepackage{mathtools}
\usepackage{booktabs} 
\usepackage{soul}
\usepackage{tikz}
\usetikzlibrary{patterns}
\usepackage{pgfplots}
\pgfplotsset{compat=1.18}
\tikzset{font=\scriptsize}
\usepackage{colortbl}
\usepackage{array}
\usepackage{xcolor}
\usepackage{xspace}
\usepackage{multirow}
\usepackage{subfig}

\definecolor{leftgreen}{RGB}{158, 210, 190} 
\definecolor{midgreen}{RGB}{126, 170, 146}
\definecolor{rightred}{RGB}{255, 217, 183}

\definecolor{napiergreen}{rgb}{0.16, 0.5, 0.0}
\definecolor{parisgreen}{rgb}{0.31, 0.78, 0.47}

\definecolor{fluorescentorange}{rgb}{1.0, 0.75, 0.0}
\definecolor{orange(colorwheel)}{rgb}{1.0, 0.5, 0.0}

\definecolor{Blueback}{RGB}{218, 227, 243} 
\definecolor{Greenback}{RGB}{226, 240, 217}
\definecolor{Redback}{RGB}{251, 229, 214} 

\newcommand{\modelname}{\textsc{MetaCritique}\xspace}

\usepackage[T1]{fontenc}

\usepackage[utf8]{inputenc}

\usepackage{microtype}

\newcommand*{\belowrulesepcolor}[1]{%
  \noalign{%
    \kern-\belowrulesep
    \begingroup
      \color{#1}%
      \hrule height\belowrulesep
    \endgroup
  }%
}
\newcommand*{\aboverulesepcolor}[1]{%
  \noalign{%
    \begingroup
      \color{#1}%
      \hrule height\aboverulesep
    \endgroup
    \kern-\aboverulesep
  }%
}

\newcommand*\samethanks[1][\value{footnote}]{\footnotemark[#1]}

\title{The Critique of Critique}

\author{Shichao Sun\textsuperscript{\rm 1,5}, Junlong Li\textsuperscript{\rm 2,5}, Weizhe  Yuan\textsuperscript{\rm 4}, Ruifeng Yuan\textsuperscript{\rm 1}, Wenjie Li\textsuperscript{\rm 1}\thanks{\,\, Corresponding Authors. \\ \indent\indent   Work done while visiting GAIR Lab.}, Pengfei Liu\textsuperscript{\rm 2,3,5}\samethanks \\
   \textsuperscript{\rm 1} The Hong Kong Polytechnic University, \textsuperscript{\rm 2} Shanghai Jiao Tong University \\ 
   \textsuperscript{\rm 3} Shanghai Artificial Intelligence Laboratory,
   \textsuperscript{\rm 4} New York University   \\
    \textsuperscript{\rm 5} Generative AI Research Lab (GAIR)\\
  {\normalsize \texttt{cswjli@comp.polyu.edu.hk}, \texttt{pengfei@sjtu.edu.cn}}
  }

\begin{document}
\maketitle
\begin{abstract}
Critique, as a natural language description for assessing the quality of model-generated content, has played a vital role in the training, evaluation, and refinement of LLMs. However, a systematic method to evaluate the quality of critique is lacking. In this paper, we pioneer the critique of critique, termed \modelname, which builds specific quantification criteria. 
To achieve a reliable evaluation outcome, we propose Atomic Information Units (AIUs), which describe the critique in a more fine-grained manner. \modelname aggregates each AIU's judgment for the overall score. Moreover, \modelname delivers a natural language rationale for the intricate reasoning within each judgment.
Lastly, we construct a meta-evaluation dataset covering 4 tasks across 16 public datasets involving human-written and LLM-generated critiques. 
Experiments demonstrate that \modelname can achieve near-human performance. Our study can facilitate future research in LLM critiques based on our following observations and released resources: (1) superior critiques judged by \modelname can lead to better refinements, indicating that it can potentially enhance the alignment of existing LLMs; (2) the leaderboard of critique models reveals that open-source critique models commonly suffer from factuality issues; (3) relevant code and data are publicly available at \url{https://github.com/GAIR-NLP/MetaCritique} to support deeper exploration; (4) an \textbf{API} at PyPI with the usage documentation in Appendix \ref{sec:api_usage} allows users to assess the critique conveniently.
\end{abstract}

\section{Introduction}
Natural language critique has assumed a crucial role in advancing the development of Large Language Models (LLMs), ranging from
the training of a more helpful and harmless model \citep{bai2022constitutional,OpenAI_GPT4_2023,scheurer2023training,wu2023fine},  alignment evaluation of model generations \citep{wang2023pandalm, zheng2023judging, chan2023chateval,li2023generative} to the refinement of defective model outputs \citep{madaan2023self, gou2023critic, selfee2023,akyurek-etal-2023-rl4f}. 
While a bunch of recent works are being done using generated critiques to assist in the development of LLMs~\citep{cui2023ultrafeedback,kim2023prometheus,wang2023shepherd,li2023generative,ke2023critiquellm}, 
there has not been enough emphasis on how to automatically and efficiently evaluate the quality of these critiques due to the following challenges:
(i) \textbf{quantification}: establishing specific criteria to qualify the critique rating, 
(ii) \textbf{reliability}: ensuring transparency to calculate the comparable score,
and (iii) \textbf{intricacy}: grasping the complex relations among multiple concepts in the critique evaluation.

\begin{figure*}[ht]
    \centering
    \includegraphics[width=0.88\linewidth]{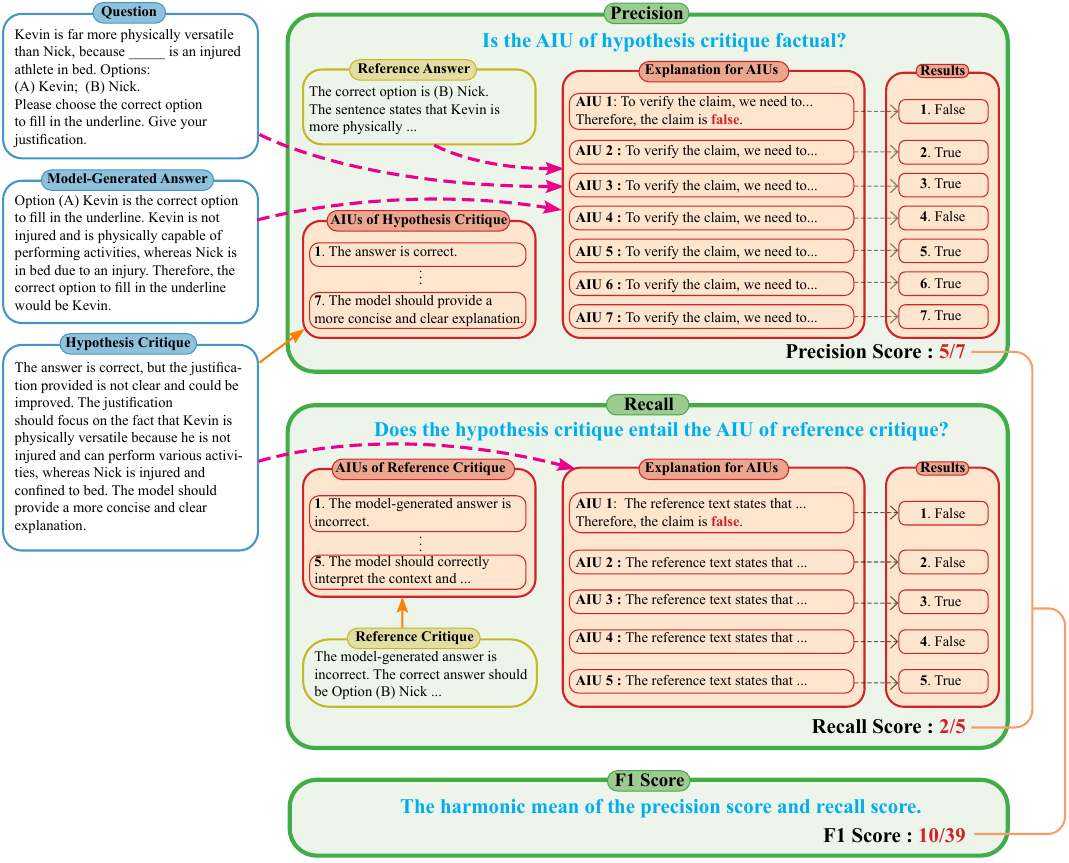}
    \caption{An example of \modelname for hypothesis critique evaluation. More details can be found in Figure \ref{tab:com-case-intro}. Atomic Information Units (AIUs) are  fundamental segments of informative critique that cannot be divided further. The ratio of ``True'' AIUs is calculated as the corresponding score. }
    \label{tab:case-intro}
\end{figure*}
In this paper, we pioneer the critique of critique, termed \modelname, to get over the above hurdles. An example of \modelname is shown in Figure \ref{tab:case-intro}. Firstly, \modelname tackles the \textbf{quantification} issue by establishing specific criteria, i.e., a meaningful critique should provide factual statements and comprehensive assessments, which will be quantified by two metrics: \textit{precision} and \textit{recall}. Precision serves to gauge the accuracy of the critique's content, ensuring each point is factual, while recall measures the extent to which the critique fully covers the necessary breadth of information, reflecting its comprehensiveness. 
Secondly, \modelname addresses the \textbf{reliability} concern by introducing Atomic Information Units (AIUs). AIUs symbolize the fundamental segments of informative critique that cannot be divided further. \modelname converts the critique-level evaluation into AIU-level evaluation to minimize ambiguity in the evaluation process. Subsequently, \modelname aggregates these AIU-level outcomes to produce the overall score, thereby guaranteeing the transparency of the scoring process. 
Lastly, inspired by the fact that complex reasoning problems can be alleviated by using LLMs with Chain-of-Thought (CoT) \citep{wei2022chain}, we attempt to resolve the \textbf{intricacy} problem by generating a natural language rationale step by step for each AIU-level judgment. As a result, it can enhance the reliability of judgment and facilitate human involvement in the evaluation loop. 

Given the absence of a dataset for critique evaluation, we curate a meta-evaluation dataset covering 4 tasks (question answering, reasoning, entailment, and summarization) across 16 public datasets, involving human-written and LLM-generated critiques. Our \modelname achieves near-human performance, indicating that \modelname can help understand human annotation and LLM's reflection. Besides, \modelname can identify high-quality critiques, which lead to improved results via iterative refinement. This indicates that \modelname can enhance the alignment of existing LLMs. A leaderboard of critique models also aids in identifying the pros and cons of various critique models. In conclusion, \modelname can potentially advance the progress of LLMs.

\begin{figure*}[ht]
    \centering
    \includegraphics[width=0.93\linewidth]{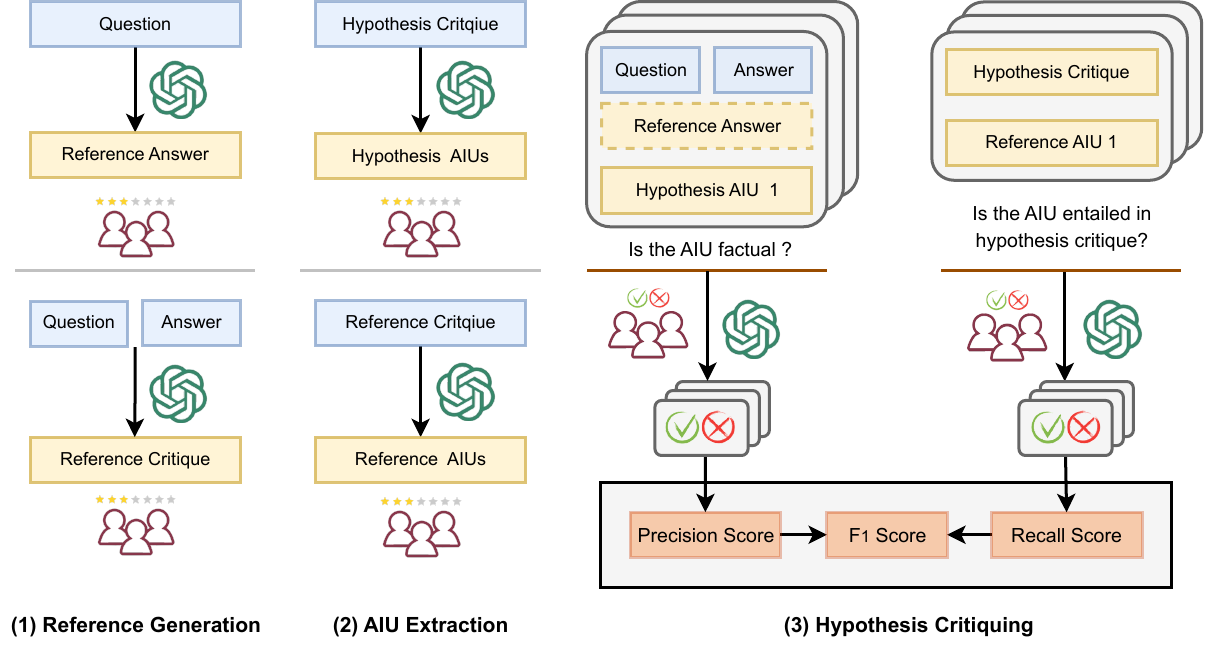}
    \caption{An overview for our \modelname powered by GPT-4 and human annotation for each step. The ``answer'' in the blue box is the model-generated answer. The star rating refers to annotators giving 1-7 likert score, and the check-and-cross mark indicates that annotators give the True or False label, just like GPT-4.}
    \label{fig:pipelines}
\end{figure*}

\section{\modelname}
\modelname evaluates the quality of a hypothesis critique by generating its critique, i.e., the critique of critique. It involves three steps: (1) reference generation, (2) AIU extraction, and (3) hypothesis critiquing as illustrated in Figure \ref{fig:pipelines}.

To facilitate the detailed description of \modelname, we first introduce some important concepts.
\noindent\textbf{Question} denotes a user's query or instruction that prompts LLMs to produce a pertinent and insightful response.
\noindent\textbf{Model-generated Answer} describes the textual content created by LLMs as a reaction to the question.
\noindent\textbf{Reference Answer} is a ground-truth answer for the question. 
\noindent\textbf{Hypothesis Critique} is the natural language feedback to point out errors of the model-generated answer and provide actionable suggestions. It can be written by either human annotators or LLMs. 
\noindent\textbf{Reference Critique} is an ideal critique to comment accurately and thoroughly on the model-generated answers. 
\noindent\textbf{Atomic Information Unit (AIU)} is the smallest unit that can self-sufficiently convey a piece of information. 
It can help reduce the ambiguity of the evaluation process and improve the transparency of the evaluation outcome.
Generating a numeric score for the critique depends on the information volume it encompasses. For example, a precision score considers the volume of correct information in the critique. However, information, being intangible, renders the task of determining the volume of information unfeasible \citep{porat1977information,soofi1994capturing}.  By extracting and counting AIUs from a critique, we can approximate the volume of information as the number of AIUs because AIUs are the most fundamental elements of critiques. 

\subsection{Step 1: Reference Generation}
\label{sec:ref_gen}
Reference is essential for most text generation evaluation \citep{papineni-etal-2002-bleu,lin-2004-rouge,Zhang2020BERTScore,yuan2021bartscore}. Also, it is crucial for our \modelname. 
We need reference answers to calculate the precision score, and reference critiques to calculate the recall score. 
However, reference answers or critiques are always unavailable because they require significant human effort. 
To address this issue, we adopt the GPT-4 generated content as a proxy of the references as in previous works \citep{OpenAI_GPT4_2023,zheng2023judging,peng2023instruction,cui2023ultrafeedback,li2023generative}. 
Moreover, we conduct human evaluations to ensure the quality of the GPT-4 generated content.
We provide detailed prompting instructions in Table \ref{tab:prompt_reference_critique}. 

\subsection{Step 2: AIU Extraction}
\label{sec:aiu_ext}
AIU extraction aims to split a critique into AIUs. It is similar to some prior works such as verifiable claims extraction for factuality detection \citep{chern2023factool} and atomic content units extraction for summarization evaluation \citep{liu2023learning}.
It has been proven that suitably prompted GPT-4 can precisely conduct such a task as mentioned in \citet{chern2023factool}. Inspired by their success, we implement AIUs extraction by prompting GPT-4.
Besides, we conduct human evaluations to verify the quality of the GPT-4 extraction.
We provide detailed prompting instructions in Table \ref{tab:prompt_aius_extraction}. 

\subsection{Step 3: Hypothesis Critiquing} 

\paragraph{Precision} We devise \textit{ precision} to verify the factuality of the hypothesis critique. 
It is motivated by the fact that good critiques should state factual information without any hallucination. 
Specifically, we design the \textbf{precision task}, which is a binary classification task on the AIU level to validate whether each AIU is factual or non-factual.
This task receives the question, model-generated answer and the optional reference answer as the context,
and outputs whether an AIU from the hypothesis critique is factual or non-factual along with a natural language rationale.

We prompt LLMs with strong instruction-following capability to implement this evaluation. 
We follow the idea of CoT reasoning to design the instructions. Firstly, the LLM needs to find the necessary information to verify the AIU. Then, the LLM explains and reasons whether the AIU is factual or not. Finally, the LLM states the conclusion. Detailed prompting instructions with demonstrations are shown in Table \ref{tab:prompt_precision}.

After checking each AIU in the hypothesis critique, we denote the precision score $s_p$ as the proportion of factual AIUs relative to the total count of AIUs in the hypothesis critique.

\paragraph{Recall}
We use \textit{recall} to assess the coverage of the hypothesis critique over the reference critique. It is motivated by the fact that good critiques should contain all key points of the reference critique without any omissions. In this evaluation, we design \textbf{recall task}, which is a binary classification task on the AIU level to classify whether the hypothesis critique entails each AIU of the reference critique.
This task receives the hypothesis critique as the premise and outputs whether an AIU from the reference critique is entailed in the hypothesis critique or not along with a natural language rationale.

We prompt LLMs guided by the CoT reasoning to perform this evaluation. Firstly, the LLM analyses if the AIU from the reference critique is mentioned or logically inferred from the hypothesis critique. Subsequently, the LLM states whether the AIU is entailed or not. Detailed prompting instructions with demonstrations are in Table \ref{tab:prompt_recall}.

After checking each AIU in the reference critique, we denote the recall score $s_r$ of the hypothesis critique as the ratio of entailed AIUs to all AIUs from the reference critique.

\paragraph{F{\small 1} Score} We introduce the F{\small 1} Score $s_f$ as an overall assessment score, which harmonizes the \textit{precision} score $s_p$ and \textit{recall} score $s_f$ as follows:
\begin{equation}
s_f = 2\frac{s_p \cdot s_r}{s_p + s_r}
\label{eq:count_f}
\end{equation}

\section{Meta-Evaluation Dataset}
\label{sec:data}
In this section, we elaborate on constructing a meta-evaluation dataset and human annotation for evaluating the critique evaluation, along with presenting its statistical features. We first collect questions, model-generated answers containing flaws, and human-written or LLM-generated critiques (hypothesis critique). Subsequently, we use GPT-4 to generate critiques and answers as references and extract AIUs from critiques. Finally, human annotators use these data to complete precision and recall tasks for each AIUs as shown in Figure \ref{fig:pipelines}.

\subsection{Collection of Question and Model-Generated Answer}

To get broad coverage of NLP domains, we collect ready-made question-answer pairs from the Shepherd \citep{wang2023shepherd} dataset, which consists of question-answer-critique triads for generating critique across various domains. The model-generated answer in this dataset has reasonable errors, so that the critique can be generated to improve the response. We carefully extract some data to exclude the tasks that need specific tools to find errors, like code generation. As a result, we collect data from four domains: entailment, reasoning, question answering, and summarization, across 16 datasets: Entailment Bank \citep{dalvi-etal-2021-explaining}, e-SNLI \citep{camburu2018snli}, Adversarial NLI \citep{nie-etal-2020-adversarial}, ECQA \citep{aggarwal-etal-2021-explanations}, CosmosQA \citep{huang-etal-2019-cosmos}, HellaSwag \citep{zellers-etal-2019-hellaswag}, WinoGrande \citep{sakaguchi2021winogrande}, PIQA \citep{bisk2020piqa}, ARC \citep{clark2018think}, RACE \citep{lai-etal-2017-race}, SIQA \citep{sap-etal-2019-social}, TriviaQA \citep{joshi-etal-2017-triviaqa}, Natural Question \citep{kwiatkowski-etal-2019-natural}, BoolQ \citep{clark-etal-2019-boolq}, GPT-3 summarization \citep{goyal2022news}, DeFacto \citep{liu-etal-2023-improving}. 

\subsection{Collection of Hypothesis Critique}

Hypothesis critique can be written by human annotators (abbreviated to \textit{\textbf{Hypo.h}} in tables) and LLMs (abbreviated to \textit{\textbf{Hypo.l}} in tables).
It is valuable to evaluate human-written critique, as it can help to understand the common shortcomings of human annotations. Also, it is necessary to understand how well the LLM-generated critiques perform because LLM-generated critiques have been widely used for training, evaluation, and refinement. 

For human-written critique, we use one ready-made critique from the Shepherd dataset. 
For LLM-generated critiques, we generate two critiques for each question-answer pair. These two critiques will be further used to conduct pairwise comparisons. We do not conduct the pairwise comparison between human-written critiques and LLM-generated critiques to avoid potentially misleading outcomes that could arise from their writing styles. 
In detail, we randomly select two LLMs from WizardLM \citep{xu2023wizardlm} (13B and 70B), GPT-3.5, LLaMA-2 \citep{touvron2023llama} (chat-13B and chat-70B), Vicuna \citep{vicuna2023} (13B and 33B), and SelFee \citep{selfee2023} (13B). Then, we respectively prompt the selected LLM to generate one critique.
We use the same prompt as the reference critique generation (Table \ref{tab:prompt_reference_critique}) for GPT-3.5. As for other LLMs, the prompt is shown in Table \ref{tab:prompt_hypothesis_critique}.

\subsection{Collection of Human Annotation}
We collect human labeling results for precision and recall tasks driven by two objectives: (1) to evaluate the performance of different LLMs to execute the precision and recall tasks (Exp. II). (2) to meta-evaluate \modelname and its counterparts in the light of human judgments (Exp. III-IV). 

Before annotation, we prepare the reference answer, reference critique and extracted AIUs via GPT-4 as introduced in Section \ref{sec:ref_gen} and \ref{sec:aiu_ext}. We engaged two postgraduate students to perform the precision and recall task. As shown in Figure \ref{fig:pipelines}, they replace the GPT-4 to provide solely binary labels without explanation. A third postgraduate meticulously reviews the work of the first two annotators, ensuring accuracy and resolving any discrepancies. This process is essential for maintaining the reliability of our research data. Lastly, we can calculate the \modelname scores via these annotated results. These scores (shown in Table \ref{tab:humanscore}) represent human judgments (named as \textbf{gold scores}).

\begin{table}[t]
  \centering
  \small
   \setlength{\tabcolsep}{6pt}
  \begin{tabular}{lcccc}
\toprule
\textbf{Type}             & \textbf{\# Pair} & \textbf{\# Critique}  & \textbf{Avg. \# AIUs} \\
\midrule
Hypo.h            & 100              & 100       & 3.31  \\
Hypo.l      & 100              & 200       & 8.10\\
Reference & 100              & 100      & 7.02\\
\bottomrule
\end{tabular}
  \caption{Statistics of meta-evaluation dataset.}
  \label{tab:stat}
\end{table}

\subsection{Statistics}
\label{sec:data_stat}
Table \ref{tab:stat} shows the statistics of our meta-evaluation dataset. We collect 100 question-answer pairs. Each pair has 1 human-written critique, 2 LLM-generated critiques, and 1 reference critique. 
We find that the number of AIUs in human-written critiques is less than half of it in LLM-generated critiques or reference critiques. This implies that human-written critiques possibly contain less information than LLM-generated critiques.

\section{Experiments}
In this section, we introduce various baselines and experiments to show the feasibility (Exp. I-II) and effectiveness (Exp. III-V) of \modelname. 

\subsection{Baseline}
We compare multiple modern LLMs for AIU-level precision and recall tasks (Exp. II). The tested models include \textbf{Zephyr} \citep{tunstall2023zephyr}, \textbf{WizardLM}, \textbf{LLaMA-2 Chat}, \textbf{Vicuna}, \textbf{GPT 3.5}, and \textbf{GPT-4}. Moreover, we randomly choose 100 AIUs and engage a postgraduate student (not the annotator) to perform the same task. This result can approximate a ceil performance (named as \textbf{Human}). We also introduce \textbf{GPT-4 w/o ans}, where we generate the reference answer and perform the precision task in one step. It aims to confirm the importance of pre-generating a reference answer.

We compare four variants of \modelname with two GPT-4 based methods introduced by \citet{wang2023shepherd} for Exp. III-V:
\noindent\textbf{MetaCritique$_{\text{GPT4}}$-P}, \textbf{MetaCritique$_{\text{GPT4}}$-R} and \textbf{MetaCritique$_{\text{GPT4}}$-F{\small 1}} is respectively the precision score, recall score and F{\small 1} score of \modelname powered by GPT-4.  
\noindent\textbf{MetaCritique$_{\text{Open}}$-F{\small 1}} is the F{\small 1} score of \modelname powered by open-source LLMs, whereby WizardLM 70B is used for the precision task, and WizardLM 13B is used for the recall task because they beat other open-source LLMs in the Table \ref{tab:generatescore}.
\noindent\textbf{Pairwise$_{\text{GPT4}}$} is to compare two hypothesis critiques via GPT-4 and pick up the better one. The prompting instruction is shown in Table \ref{tab:pairwise_gpt_prompt}. 
\noindent\textbf{Single$_{\text{GPT4}}$} is to generate a likert score (1-7) for a hypothesis critique via GPT-4. The prompting instruction is shown in Table \ref{tab:single_gpt_prompt}. 

\begin{table}[t]
\centering
\small
\begin{tabular}{ccccc}
\toprule
\multicolumn{2}{c}{\textbf{Reference Generation}} & \multicolumn{3}{c}{\textbf{AIUs Extraction}} \\ \cmidrule(lr){1-2} \cmidrule(lr){3-5}
 Answer             & Critique            & Hypo.h    & Hypo.l    & Reference   \\ \midrule
 6.51               & 6.56                & 6.72      & 6.57      & 6.79 \\
\bottomrule
\end{tabular}
\caption{Human evaluation for GPT-4 outcomes (Likert score on 1-7 scale). 1 is the worst, and 7 is the best.}
\label{tab:humaneval}
\end{table}

\subsection{Exp-I: Human Evaluation for Reference Generation and AIUs Extraction}
\textbf{Q1: Can GPT-4 outcomes serve as references?}
\paragraph{Setup}
We conduct a human evaluation to validate the quality of the reference answer, reference critique, and extracted AIUs that are generated by GPT-4. We engaged two postgraduate students as annotators. We ask each human annotator to rate the outcome on a 1–7 likert score. Detailed instructions for human annotators to evaluate reference answer generation, reference critique generation, and AIUs extraction can be found in Table \ref{tab:human_eval_answer}, \ref{tab:human_eval_critique}, and \ref{tab:human_eval_aiu}, respectively. To obtain a reliable evaluation outcome, we present as much information as possible to annotators. We also allow annotators to carefully search online whenever they need help. 
\paragraph{Results}
Table \ref{tab:humaneval} shows the rating scores for the quality of GPT-4 generated answers, critiques, and AIUs. The left half shows that GPT-4 attains remarkable performance, which can confirm the feasibility of using GPT-4 generated outcomes as references. Moreover, the right half shows that GPT-4 delivers impressive results, which can justify its effective use to extract AIUs for \modelname. 

\begin{table}[t]
\centering
\small
\setlength{\tabcolsep}{3.5pt}
\begin{tabular}{lccccc}
\toprule
      &     & \multicolumn{2}{c}{\textbf{Precision Task}} & \multicolumn{2}{c}{\textbf{Recall Task}} \\ \cmidrule(lr){3-4} \cmidrule(lr){5-6} 
   Model  &  Size   & Hypo.h         & Hypo.l          & Hypo.h        & Hypo.l        \\ \midrule
Human &  -- & 90.00        & 86.00        & 86.00        & 85.00       \\  \midrule
\multirow{3}{*}{Llama 2 Chat} & 7B  &    31.42      &  34.14     &  8.55  &    5.98   \\
 & 13B  & 56.50         & 52.22          & 55.13      & 51.78       \\
 & 70B & 54.08     & 58.09          & 72.08       & 65.46        \\ \midrule
\multirow{3}{*}{Vicuna}  & 7B & 71.00        & 70.37        & 59.69      & 60.83     \\
  & 13B & 72.81        & 69.14        & 71.94      & 70.66      \\
  & 33B & 62.84        & 60.12         & 72.93      & 63.82     \\ \midrule
Zephyr-$\beta$   & 7B  & 62.84       & 59.44        & 54.56     & 61.82 \\ \midrule
\multirow{4}{*}{WizardLM}   & 7B  & 26.89       & 30.19         & 52.28     & 55.56   \\ 
 & 13B  & 62.54       & 59.81        & \underline{74.50}      & \underline{74.22}     \\ 
  & 30B & 56.50        & 63.89         & 63.53     & 61.68      \\
  & 70B & \underline{79.76}       & \underline{72.28}       & 72.51    & 71.44     \\ \midrule
GPT-3.5 & -- & 81.87         & 77.28         & 80.63         & 81.41        \\
GPT-4 w/o ans  & -- & 86.40       &  81.48      & --       & -- \\
GPT-4   & -- & \textbf{89.12}        &\textbf{87.96}        & \textbf{85.47}        & \textbf{86.82}   \\
\bottomrule    
\end{tabular}
  \caption{AIU-level accuracy. \underline{Underline} is the best result among all open-source LLMs, and \textbf{bold} is the best outcome among all LLMs.}
  \label{tab:generatescore}
\end{table}

\subsection{Exp-II: AIU-level Accuracy}
\textbf{Q2: Which LLMs are capable of powering \modelname?}
\paragraph{Setup} 
Our \modelname centers around two binary classification tasks: precision and recall. In this experiment, we investigated the capability of various LLMs to execute two tasks. Each AIU was treated as a unique test case. 

\paragraph{Results}
We present the AIU-level accuracy results in Table \ref{tab:generatescore}. We find that GPT-4 outperforms all LLMs by a large margin and achieves an impressive performance (\textbf{nearly 90\%}), which is comparable to that of humans. This shows that it is reasonable to use \modelname powered by GPT-4 for automated evaluation of human-written critiques and LLM-generated critiques. 
Moreover, WizardLM-70B and WizardLM-13B stand out as the top-performing open-source models in precision tasks and recall tasks, respectively. Remarkably, they rival closely  GPT-3.5. Lastly, the degradation (around 3\% and 6\%) without the reference answer indicates that it is necessary to pre-generate a reference answer.

\begin{table*}[t]
\centering
\small
\setlength{\tabcolsep}{15pt}
\begin{tabular}{lcccccc}
\toprule
   \multirow{2}{*}{\textbf{Methods}}       & \multicolumn{2}{c}{\textbf{Pearson}} & \multicolumn{2}{c}{\textbf{Spearman}} & \multicolumn{2}{c}{\textbf{Kendall's Tau}} \\ \cmidrule(lr){2-3} \cmidrule(lr){4-5} \cmidrule(lr){6-7}
          & Hypo.h          & Hypo.l          & Hypo.h         & Hypo.l      & Hypo.h         & Hypo.l        \\ \midrule
Single$_{\text{GPT4}}$   & 0.508       & 0.390         & 0.503    & 0.379  & 0.396     & 0.290     \\ \midrule
MetaCritique$_{\text{Open}}$-F{\small 1} & 0.730$\star$         & 0.709$\star$        & 0.745$\star$       & 0.690$\star$  & 0.566$\star$        & 0.506$\star$       \\
MetaCritique$_{\text{GPT4}}$-P & 0.355         & 0.667$\star$        & 0.283       & 0.681$\star$  & 0.227        & 0.504$\star$       \\
MetaCritique$_{\text{GPT4}}$-R   & \textbf{0.844}$\dag$        & 0.831$\star$        & \textbf{0.837}$\dag$      & 0.830$\star$   & \textbf{0.681}$\dag$      & 0.649$\star$  \\
MetaCritique$_{\text{GPT4}}$-F{\small 1}   & 0.841$\star$         &  \textbf{0.886}$\dag$  & 0.836$\star$        &\textbf{0.899}$\dag$  & 0.675$\star$        & \textbf{0.724 }$\dag$\\
\bottomrule    
\end{tabular}
  \caption{Correlation between different models with the gold scores. $\star$ means significantly ($p<0.05$) outperforms the baseline method (Single$_{\text{GPT4}}$). $\dag$means significantly ($p<0.05$) outperforms all methods. }
  \label{tab:correlation}
\end{table*}

\subsection{Exp-III: Correlation Coefficient}
\textbf{Q3: Which evaluation methods can give rating scores that are close to human judgments?}
\paragraph{Setup}
In this experiment, we use different methods to generate a score for the hypothesis critique. We calculate correlation coefficients to measure the correlation between different scoring baselines and human judgments. Specifically, we use Pearson correlation \citep{lee1988thirteen,mukaka2012guide}, Spearman correlation \citep{zar2005spearman}, and Kendall's Tau \citep{kendall1938new} as metrics. We calculate the above correlation coefficients between the outcome score and the gold F{\small 1} score. To perform a rigorous analysis, we adopt the bootstrapping method \citep{koehn-2004-statistical} for significance tests.  

\paragraph{Results}
In Table \ref{tab:correlation}, we show the correlation between various methods with human judgments. Our MetaCritique$_{\text{GPT4}}$-F{\small 1} beats the Single$_{\text{GPT4}}$ baseline by a large margin, confirming its increased reliability. We also observe that MetaCritique$_{\text{GPT4}}$-P has a reduced correlation in human-written critiques, likely because humans make fewer factuality mistakes, resulting in clustered precision scores. In addition, MetaCritique$_{\text{Open}}$ distinctly exceeds the performance of the Single$_{\text{GPT4}}$ baseline, demonstrating that even less advanced LLMs can surpass Single$_{\text{GPT4}}$ baseline. This indicates that our \modelname framework is more effective than simple GPT-4 scoring. Lastly, precision and recall scores complement each other in assessing LLM-generated critiques. MetaCritique$_{\text{GPT4}}$-P or MetaCritique$_{\text{GPT4}}$-R is slightly inferior to MetaCritique$_{\text{GPT4}}$-F{\small 1}.

\subsection{Exp-IV: Pairwise Comparison}
\textbf{Q4: Which evaluation methods can choose the critiques that humans prefer?}
\paragraph{Setup}
In this experiment, we utilize a range of scoring baselines to identify the better critique out of two critiques generated by LLMs. 
We calculate the agreement rate to evaluate the performance, defined as the consistency of the superior critique chosen by various methods with the gold standard critique determined by gold F{\small 1} score. 

\begin{figure}[t]
\centering
\begin{tikzpicture}
    \begin{axis}
        [xbar,
        legend image code/.code={%
                        \draw[#1] (0cm,-0.07cm) rectangle (0.2cm, 0.13cm);
        },  
        reverse legend,
        tick label style={font=\scriptsize},
        legend style={font=\scriptsize},
        label style={font=\scriptsize},
        ytick={0, 1, 2, 3, 4, 5},
        yticklabels={MetaCritique$_{\text{GPT4}}$-F1, MetaCritique$_{\text{GPT4}}$-P, MetaCritique$_{\text{GPT4}}$-R, MetaCritique$_{\text{Open}}$-F1, Pairwise$_{\text{GPT4}}$, Single$_{\text{GPT4}}$},
        bar width= 9pt,
        xmin=20,
        xmax=90,
        xtick={20, 40, 60, 80},
        xticklabels={20\%, 40\%, 60\%, 80\%},
        width=0.38\textwidth,
        height=0.27\textwidth,
        bar shift=0pt,
        nodes near coords, 
        nodes near coords align={horizontal}, 
        nodes near coords style={
                anchor=west,
        },
        enlarge y limits={abs=0.9},
        axis y line*=none,
        axis x line*=bottom,
        point meta=explicit symbolic,
        ] 
        \addplot [draw = orange(colorwheel),
        semithick,
        fill = white,
        postaction = {
            pattern = crosshatch,
            pattern color = fluorescentorange,
            },
        ] coordinates {
            (77,0) [77\%]
        }; 
        \addplot [draw = orange(colorwheel),
        semithick,
        fill = white,
        postaction = {
            pattern = crosshatch,
            pattern color = fluorescentorange,
            },
        ] coordinates {
            (68,1) [68\%]
        }; 
        \addplot [draw = orange(colorwheel),
        semithick,
        fill = white,
        postaction = {
            pattern = crosshatch,
            pattern color = fluorescentorange,
            },
        ] coordinates {
            (47,2) [47\%]
        }; 
                \addplot [draw = orange(colorwheel),
        semithick,
        fill = white,
        postaction = {
            pattern = crosshatch,
            pattern color = fluorescentorange,
            },
        ] coordinates {
            (69,3) [69\%]
        }; 
        \addplot [draw = napiergreen,
        semithick,
        fill = white,
        postaction = {
            pattern = north east lines,
            pattern color = parisgreen,
            },
        ] coordinates {
            (55,4) [55\%]
        }; 
        \addplot [draw = napiergreen,
        semithick,
        fill = white,
        postaction = {
            pattern = north east lines,
            pattern color = parisgreen,
            },
        ] coordinates {
            (33,5) [33\%]
        }; 
    \end{axis} 
\end{tikzpicture}
\caption{Agreement rate for pairwise comparison.}
\label{fig:pairwise-agreement}
\end{figure}
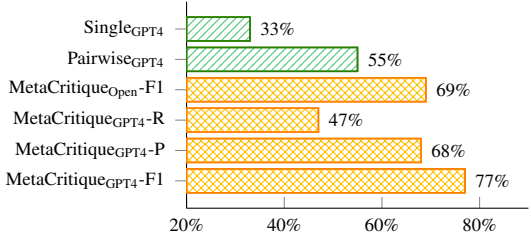

\paragraph{Results}
Figure \ref{fig:pairwise-agreement} illustrates the agreement rate of various methods.
Our MetaCritique$_{\text{GPT4}}$-F{\small 1} achieves the best performance. Especially, it exceeds the Single$_{\text{GPT4}}$ baseline by a considerable margin (\textbf{44\%}). It implies that the scores of MetaCritique$_{\text{GPT4}}$-F{\small 1} are more comparable than Single$_{\text{GPT4}}$ baseline. Even MetaCritique$_{\text{Open}}$ significantly outperforms GPT-4 powered baselines Single$_{\text{GPT4}}$ and Pairwise$_{\text{GPT4}}$, confirming the effectiveness of our framework. Finally, precision and recall scores serve as complementary metrics for evaluating critiques, because MetaCritique$_{\text{GPT4}}$-P and MetaCritique$_{\text{GPT4}}$-R are somewhat less effective than MetaCritique$_{\text{GPT4}}$-F{\small 1}.

\subsection{Exp V: Better Critique, Better Refinement}
\textbf{Q5: Can critique evaluations improve the alignment of existing LLMs?}
\paragraph{Setup}
Critique is commonly applied to improve the quality of model outputs via refinement \citep{madaan2023self}. It is intuitive that superior critiques result in better refinements. To confirm this hypothesis, we conduct this experiment. Specifically, we instruct GPT-4 to refine the model outputs via the critique. Detailed instructions with demonstrations are presented in Table \ref{tab:refine_gpt_prompt}. Subsequently, we compare the refined outcomes to choose the better one. We conduct GPT-4 evaluation and human evaluation for this comparison. The prompt for GPT-4 evaluation is shown in Table \ref{tab:eval_answer_gpt}, while the equivalent instruction for human evaluators is outlined in Table \ref{tab:eval_answer_human}.

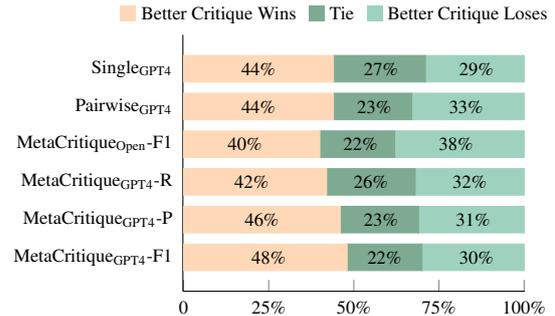
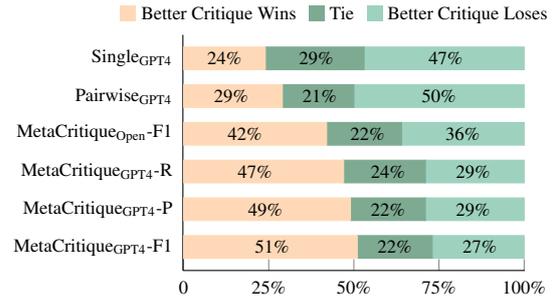
\begin{figure}[t]
\centering
\subfloat[\centering GPT-4 Evaluation]{{
\label{fig:betterRgpt}
\begin{tikzpicture}
\begin{axis}[
    xbar stacked,
    legend image code/.code={%
                    \draw[#1, draw=none] (0cm,-0.07cm) rectangle (0.2cm, 0.13cm);
    },  
    legend style={
        legend columns=3,
        at={(0.45,1)},
        anchor=south,
        draw=none,
        /tikz/every even column/.append style={column sep=0.1cm},
    },
    ytick=data,
    axis y line*=none,
    axis x line*=bottom,
    tick label style={font=\scriptsize},
    legend style={font=\scriptsize},
    label style={font=\scriptsize},
    xtick={0,25,50,75,100},
    width=.38\textwidth,
    bar width=10pt,
    yticklabels={MetaCritique$_{\text{GPT4}}$-F1, MetaCritique$_{\text{GPT4}}$-P, MetaCritique$_{\text{GPT4}}$-R, MetaCritique$_{\text{Open}}$-F1, Pairwise$_{\text{GPT4}}$, Single$_{\text{GPT4}}$},
    xticklabels={0, 25\%, 50\%, 75\%, 100\%},
    xmin=0,
    xmax=100,
    y=5mm,
    enlarge y limits={abs=0.9},
    enlarge x limits={abs=0.2},
]
\addplot[rightred,fill=rightred] coordinates
{(48,0) (46,1) (42,2) (40,3) (44,4) (44,5)};
\addplot[midgreen,fill=midgreen] coordinates
{(22,0) (23,1) (26,2) (22,3) (23,4) (27,5)};
\addplot[leftgreen,fill=leftgreen] coordinates
{(30,0) (31,1) (32,2) (38,3) (33,4) (29,5)};
\legend{Better Critique Wins, Tie, Better Critique Loses}
\coordinate (0 win) at (25,0);
\coordinate (0 tie) at (59,0);
\coordinate (0 lose) at (86,0);
\coordinate (1 win) at (23,5mm);
\coordinate (1 tie) at (58,5mm);
\coordinate (1 lose) at (85,5mm);
\coordinate (2 win) at (20,10mm);
\coordinate (2 tie) at (55,10mm);
\coordinate (2 lose) at (84,10mm);
\coordinate (3 win) at (18,15mm);
\coordinate (3 tie) at (52,15mm);
\coordinate (3 lose) at (80,15mm);
\coordinate (4 win) at (22,20mm);
\coordinate (4 tie) at (56,20mm);
\coordinate (4 lose) at (83,20mm);
\coordinate (5 win) at (22,25mm);
\coordinate (5 tie) at (58,25mm);
\coordinate (5 lose) at (86,25mm);
\end{axis} 
\node at (0 win) {48\%};
\node at (0 tie) {22\%};
\node at (0 lose) {30\%};
\node at (1 win) {46\%};
\node at (1 tie) {23\%};
\node at (1 lose) {31\%};
\node at (2 win) {42\%};
\node at (2 tie) {26\%};
\node at (2 lose) {32\%};
\node at (3 win) {40\%};
\node at (3 tie) {22\%};
\node at (3 lose) {38\%};
\node at (4 win) {44\%};
\node at (4 tie) {23\%};
\node at (4 lose) {33\%};
\node at (5 win) {44\%};
\node at (5 tie) {27\%};
\node at (5 lose) {29\%};
\end{tikzpicture}
}}%

\subfloat[\centering Human Evaluation]{{
\label{fig:betterRhuman}
\begin{tikzpicture}
\begin{axis}[
    xbar stacked,
    legend image code/.code={%
                    \draw[#1, draw=none] (0cm,-0.07cm) rectangle (0.2cm, 0.13cm);
    },  
    legend style={
        legend columns=3,
        at={(0.45,1)},
        anchor=south,
        draw=none,
        /tikz/every even column/.append style={column sep=0.1cm},
    },
    ytick=data,
    axis y line*=none,
    axis x line*=bottom,
    tick label style={font=\scriptsize},
    legend style={font=\scriptsize},
    label style={font=\scriptsize},
    xtick={0,25,50,75,100},
    width=.38\textwidth,
    bar width=3mm,
    yticklabels={MetaCritique$_{\text{GPT4}}$-F1, MetaCritique$_{\text{GPT4}}$-P, MetaCritique$_{\text{GPT4}}$-R, MetaCritique$_{\text{Open}}$-F1, Pairwise$_{\text{GPT4}}$, Single$_{\text{GPT4}}$},
    xticklabels={0, 25\%, 50\%, 75\%, 100\%},
    xmin=0,
    xmax=100,
    y=5mm,
    enlarge y limits={abs=0.625},
    enlarge x limits={abs=0.2},
]
\addplot[rightred,fill=rightred] coordinates
{(51,0) (49,1) (47,2) (42,3) (29,4) (24,5)};
\addplot[midgreen,fill=midgreen] coordinates
{(22,0) (22,1) (24,2) (22,3) (21,4) (29,5)};
\addplot[leftgreen,fill=leftgreen] coordinates
{(27,0) (29,1) (29,2) (36,3) (50,4) (47,5)};
\legend{Better Critique Wins, Tie, Better Critique Loses}
\coordinate (0 win) at (26,0);
\coordinate (0 tie) at (62,0);
\coordinate (0 lose) at (87,0);
\coordinate (1 win) at (24,5mm);
\coordinate (1 tie) at (60,5mm);
\coordinate (1 lose) at (85,5mm);
\coordinate (2 win) at (21,10mm);
\coordinate (2 tie) at (60,10mm);
\coordinate (2 lose) at (85,10mm);
\coordinate (3 win) at (18,15mm);
\coordinate (3 tie) at (54,15mm);
\coordinate (3 lose) at (82,15mm);
\coordinate (4 win) at (14,20mm);
\coordinate (4 tie) at (40,20mm);
\coordinate (4 lose) at (75,20mm);
\coordinate (5 win) at (12,25mm);
\coordinate (5 tie) at (39,25mm);
\coordinate (5 lose) at (77,25mm);
\end{axis} 
\node at (0 win) {51\%};
\node at (0 tie) {22\%};
\node at (0 lose) {27\%};
\node at (1 win) {49\%};
\node at (1 tie) {22\%};
\node at (1 lose) {29\%};
\node at (2 win) {47\%};
\node at (2 tie) {24\%};
\node at (2 lose) {29\%};
\node at (3 win) {42\%};
\node at (3 tie) {22\%};
\node at (3 lose) {36\%};
\node at (4 win) {29\%};
\node at (4 tie) {21\%};
\node at (4 lose) {50\%};
\node at (5 win) {24\%};
\node at (5 tie) {29\%};
\node at (5 lose) {47\%};
\end{tikzpicture}
}}%
\caption{Win rates of refined results from superior critique over inferior critique. The left-hand models are used to choose the better critique. A larger yellow area means a more reliable critique.}%
\label{fig:tasks} 
\end{figure}

\paragraph{Results}
We report the win rates of refined outcomes from superior critique over inferior critique, where the critique is evaluated by different methods. The outcomes of GPT-4 evaluation are depicted in Figure \ref{fig:betterRgpt}, while the results of the human evaluation appear in Figure \ref{fig:betterRhuman}. The superior critique chosen by MetaCritique$_{\text{GPT4}}$-F{\small 1} enhances the refinement significantly. Moreover, precision and recall scores are mutually supportive. Relying solely on one metric can lead to diminished performance.

\section{Exp VI: \modelname Leaderboard}
\textbf{Q6: How do various critique models perform?}

We use \modelname (MetaCritique$_{\text{GPT4}}$-F{\small 1}) to rank the critique models, such as \textbf{GPT 3.5}, \textbf{SelFee} (13B), \textbf{UltraCM} (13B), \textbf{AUTO-J} (13B), and \textbf{Human} annotators stemming from Shepherd dataset.
As shown in Table \ref{tab:leaderboard}, AUTO-J is the best open-source critique model, delivering more factual and comprehensive feedback than its open-source counterparts. 
In addition, human and GPT-3.5 achieve precision scores exceeding 80\%, outshining the performance of open-source critique models. This finding highlights that the research of open-source critique models should pay more attention to factuality issues.

\begin{table}[t]
\centering
  \small
   \setlength{\tabcolsep}{10pt}
\begin{tabular}{lccc}
\toprule
    \multirow{2}{*}{\textbf{Models}}    & \multicolumn{3}{c}{\textbf{MetaCritique$_{\text{GPT4}}$}} \\ \cmidrule(lr){2-4}
            &   \textbf{Precision}    &     \textbf{Recall}  & \textbf{F{\small 1} Score }        \\ \midrule
Human  & \textbf{83.19}       &   60.65   &  64.02   \\
GPT 3.5  & 80.79       &   64.27   &  68.72   \\  \midrule
SelFee     &  69.56     &   51.05   &  54.22   \\  
UltraCM     &    73.64     &   66.77      &  67.79 \\  
AUTO-J     &   \underline{76.43}     &    \textbf{70.65}    &  \textbf{71.14}  \\  
\bottomrule    
\end{tabular}
  \caption{\modelname scores of critique models. }
  \label{tab:leaderboard}
\end{table}

\section{Related Work}

\subsection{Critique Evaluation}
In light of the rapid advancements in LLMs, the significance of generating critique is increasingly acknowledged by researchers \citep{saunders2022self,wang2023pandalm,gou2023critic,madaan2023self,kim2023prometheus,selfee2023,wang2023shepherd,welleck2023generating,li2023generative,ke2023critiquellm,cui2023ultrafeedback}. However, the field is hindered by the scarcity of adequate research on critique evaluation.
Critique evaluation aims to evaluate the quality of the critique. It mainly relied on the human annotators \citep{saunders2022self,wang2023shepherd}, which entails considerable costs and carries a substantial risk of subjectivity. \citet{wang2023shepherd} use GPT-4 to replace human annotators, but it still lacks transparency since the numeric score is produced directly via GPT-4 without any fine-grained calculation explanation. 

\subsection{Meta Evaluation}
Meta evaluation is designed to assess automated metrics by determining the degree of correlation between automated scores and human evaluations \citep{Zhang2020BERTScore,yuan2021bartscore,sai2022survey,fu2023gptscore}. This is achieved by utilizing correlation coefficients. The Spearman correlation \citep{zar2005spearman} evaluates the monotonic relationship between two variables, focusing on their ranked values rather than raw data. Conversely, the Pearson correlation \citep{lee1988thirteen,mukaka2012guide} is used to gauge the linear relationship between two variables, employing the actual data values. Furthermore, Kendall's Tau \citep{kendall1938new} is utilized to ascertain the ordinal association between two quantified variables. Lastly, the significance test \citep{williams1959regression,koehn-2004-statistical} serves as a crucial supplementary technique to measure the improved correlations. 

\subsection{Factuality Detection}
Factuality detection aims to classify whether a textual statement, termed claim, is factual~\citep{wang-2017-liar, thorne-etal-2018-fever,augenstein-etal-2019-multifc,wadden-etal-2020-fact,guo-etal-2022-survey}.
\citet{thorne-etal-2018-fever} introduce the FEVER dataset to verify the given claim without related evidence, which leads to fact-checking models \citep{zhong-etal-2020-reasoning, krishna-etal-2022-proofver}. 
Besides, \citet{kamoi2023wice} classify whether a given claim can be entailed by the provided evidence. Similarly, in summarization tasks, FactCC \citep{kryscinski-etal-2020-evaluating} and QAGS-based models \citep{wang-etal-2020-asking} determine whether the produced summaries or summary sentences align factually with the provided document(s). 
Lastly, \citet{gao2023rarr} and \citet{chern2023factool} utilize LLMs to detect factuality without giving any claim and evidence. 

\section{Conclusion}
We are the pioneers in prioritizing critique evaluation and introducing the critique of critique, termed \modelname, involving two principles: \textit{precision} and \textit{recall}. Our \modelname is quantitative, reliable, and interpretable, wherein critiques are decomposed into AIUs, and two pragmatic tasks are established to calculate objective numeric scores along with plausible natural language rationale. We curate a meta-evaluation dataset containing various tasks to confirm the feasibility and effectiveness of \modelname. Experiments also show that superior critiques chosen by \modelname can result in a better refined outcome, suggesting that \modelname holds the potential to enhance generative artificial intelligence. Moreover, \modelname can aid in identifying the pros and cons of various critique models. This paper can facilitate future research in LLM critiques.

\section*{Acknowledgements}
We thank Ziqiang Cao, Yiyuan Li, and Yuqing Yang for their valuable feedback and suggestions on the writing of this paper. We also thank the anonymous reviewers for their valuable feedback and helpful suggestions. This project is supported by Qingyuan Research Project and Shanghai Artificial Intelligence Laboratory, National Natural Science Foundation of China (62076212), and Research Grants Council of Hong Kong (PolyU/15203617 and PolyU/5210919).

\section*{Limitations}
The creative tasks are not suitable for the recall principle, especially when there are multiple high-quality answers. It is intuitive that the suggestions for these tasks are various, so the critiques are also diverse. As a result, it is very hard to collect all effective critiques for the recall principle. 

In future work, it is essential to expand upon exploring open-source large language models to achieve more robust and dependable evaluations. This also leads to reliability and numerical stability, through the transparent and verifiable process inherent in open-source developments.

\section*{Ethical Considerations}

Our experimental data stems from Shepherd, which is well-established and publicly available. Dataset construction and annotation are consistent with the intellectual property and privacy rights of the original authors. The participants in our human evaluation were transparently informed of our research intent and were paid reasonable wages. All in all, this work complies with the ACL Ethics Policy\footnote{\url{https://www.aclweb.org/portal/content/acl-code-ethics}}. 

\bibliography{anthology,custom}
\bibliographystyle{acl_natbib}

\appendix

\section{Case Study}
\begin{figure*}[ht]
    \centering
    \includegraphics[width=\linewidth]{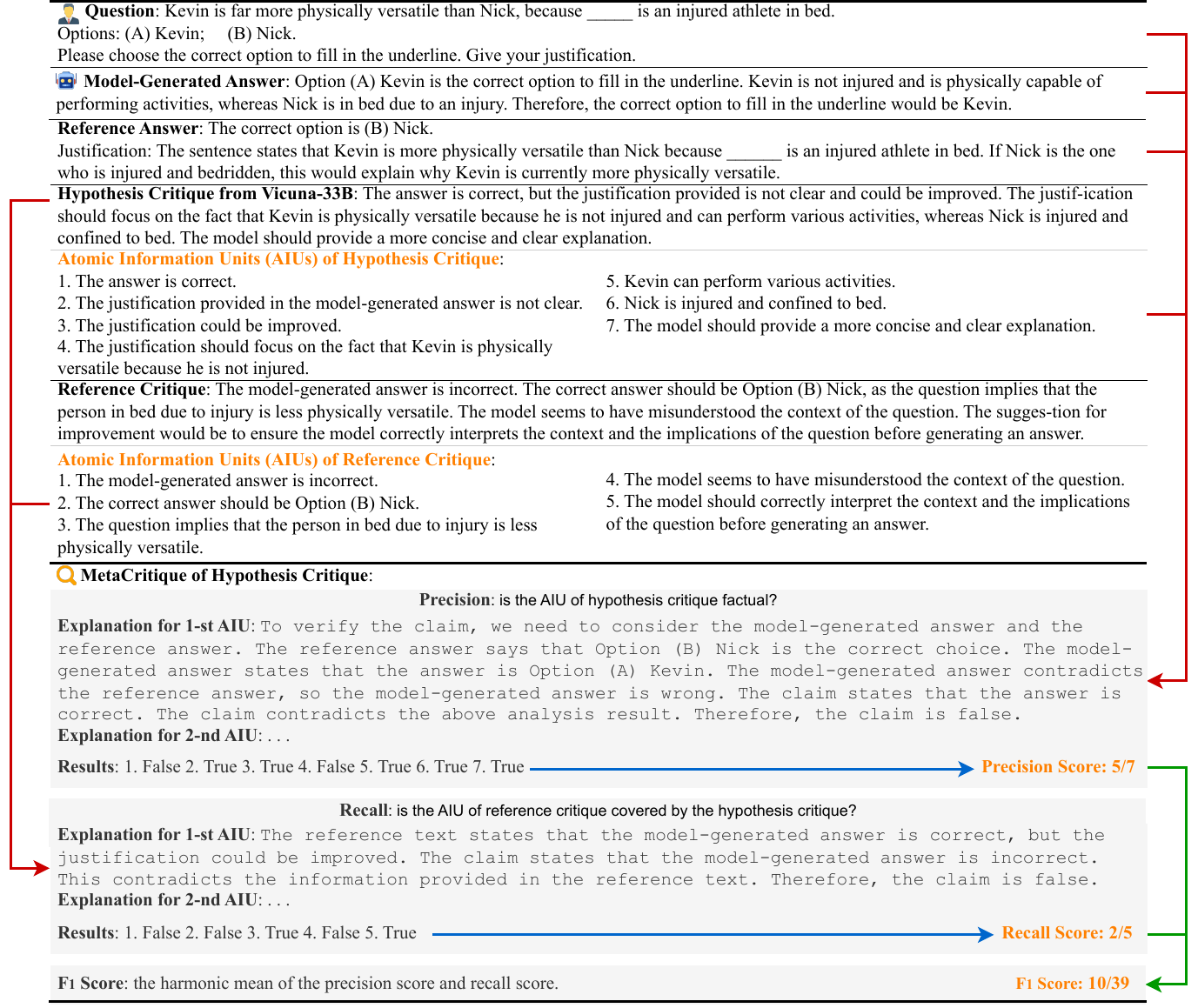}
    \caption{A case of \modelname for hypothesis critique evaluation.}
    \label{tab:com-case-intro}
\end{figure*}

We present a case in Figure \ref{tab:com-case-intro}, which contains more details of Figure \ref{tab:case-intro}, in order to illustrate how to get our \modelname scores. 

\section{Gold Score}
According to human annotation, we calculate the gold scores of precision, recall, and F{\small 1} score at the AIU level and critique level. The results are shown in Table \ref{tab:humanscore}. We find that human-written critiques get higher precision scores yet lower recall scores than LLM-generated critiques. It indicates that LLMs are even more prone to creating fictional information, while human is hardly possible to make such mistakes. Also, it is worth noting that humans miss more key information than LLMs. This might be because human-written critiques contain fewer AIUs than LLM-generated critiques.

\begin{table}[ht]
\centering
  \small
   \setlength{\tabcolsep}{8pt}
\begin{tabular}{lccc}
\toprule
          & \multicolumn{3}{c}{AIU Level (Micro)} \\ 
          &   Precision    &     Recall  & F1 Score         \\ \midrule

Hypo.h  & 87.61        &   48.72   &  62.62   \\
Hypo.l     & 71.85        & 53.28         &   61.19  \\  \midrule

& \multicolumn{3}{c}{Critique Level (Macro)} \\
& Precision   &  Recall   & F1 Score    \\ \midrule
Hypo.h  &      85.37        &     50.97    &  58.24 \\
Hypo.l     &     71.07        &   54.37 & 58.20 \\ 
\bottomrule    
\end{tabular}
  \caption{Gold scores of \modelname. }
  \label{tab:humanscore}
\end{table}

\section{API Usage at PyPI}
\label{sec:api_usage}
We publish \modelname as a Python package at PyPI in order to allow users to assess the critique conveniently. Table \ref{tab:api_usage} shows how to use \modelname in the Python code after installation. 

\begin{table}[ht]
\centering
\includegraphics[width=0.48\textwidth]{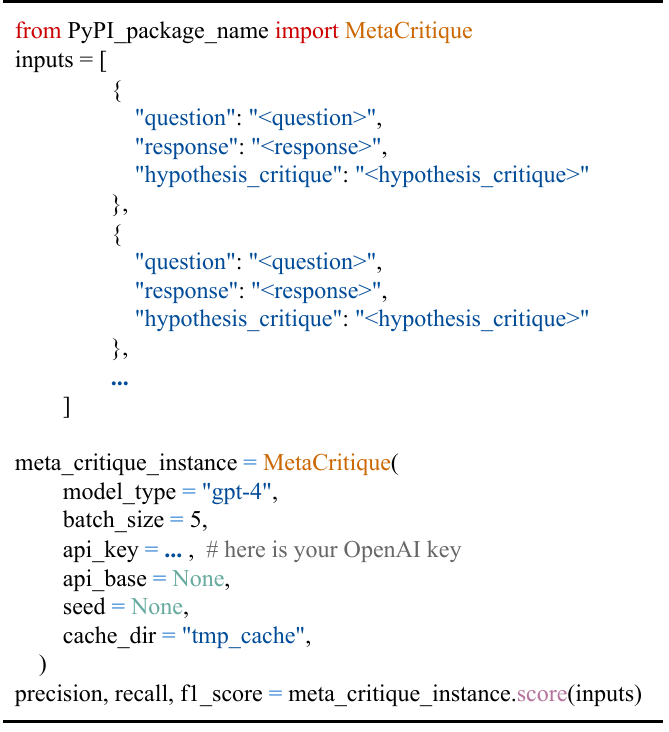}
\caption{Using \modelname in the Python code.}
\label{tab:api_usage}
\end{table}

\section{Prompts and Instructions}
\label{sec:prompts}
We elaborate the prompts for GPT-4 generation, including generating hypothesis/reference critique, extracting AIUs, performing precision/recasll tasks, refining model-generated answer, implementing GPT-4 baselines (Single$_{\text{GPT4}}$ and Pairwise$_{\text{GPT4}}$), and evaluating refined answer. Additionally, we provide guidelines for human evaluation, covering generated outcomes such as AIUs, reference answer/critique, refined answer. For creating reference answers, the system prompt employed in our paper is  ``You are a brilliant AI assistant.'' for GPT-4. In addition, we used ``claims'' instead of ``AIUs'' in our prompts to reduce possible ambiguity and confusion for LLMs.

\begin{table*}[ht]
    \scriptsize
    \centering
\begin{tabular}{@{}p{\textwidth}@{}}
\toprule
--------------SYSTEM MESSAGE-------------

~

You are a brilliant AI assistant. You will receive an input question and the model-generated answer. You need to generate the specific and actionable critiques for the model-generated answer, which contain the critical comments and helpful suggestions.

~

--------------USER MESSAGE-------------

~

input question:

\{question\}

model-generated answer:

\{model-generated answer\}

critique:
\\
\bottomrule
\end{tabular}
    \caption{Prompt for generating reference critique.}
    \label{tab:prompt_reference_critique}
\end{table*}

\begin{table*}[ht]
    \scriptsize
    \centering
\begin{tabular}{@{}p{\textwidth}@{}}
\toprule

You are a brilliant AI assistant. You will receive an input question and the model-generated answer. You need to generate the specific and actionable critiques for the model-generated answer, which contain the critical comments and helpful suggestions.

input question:

\{question\}

model-generated answer:

\{model-generated answer\}

critique:
\\
\bottomrule
\end{tabular}
    \caption{Prompt for generating hypothesis critique.}
    \label{tab:prompt_hypothesis_critique}
\end{table*}

\begin{table*}[ht]
    \scriptsize
    \centering
\begin{tabular}{@{}p{\textwidth}@{}}
\toprule
--------------SYSTEM MESSAGE-------------

~

You are a brilliant AI assistant.

You receive a critique as the input, which is the critical comment for an input question and a model-generated answer.

You need to generate all check-worthy claims of the critique.

A claim is an atomic statement that only contains a fine-grained information unit of a critique.

Each claim should be concise (a sentence) and self-contained.

Note that the 'answer' in the critique means the model-generated answer or the correct answer.

Some examples are as following:

[example 1 start]

input:

The answer violates basic principles of common sense. Flour is not sweet. Dusting it onto the bread would not sweeten the bread. Therefore, the right answer is to dust powdered sugar; sugar is, of course, sweet.

claims:

The model-generated answer violates basic principles of common sense that flour is not sweet.

Dusting Flour onto the bread would not sweeten the bread.

The correct answer is to dust powdered sugar.

Sugar is sweet.

[example 1 end]

[example 2 start]

input:

The output makes a logical error in the first bullet point of the answer, where it rejects the possibility of sunlight being the right answer. While sunlight might be produced in the sun, it doesn't stay there. Since sunlight comes to earth, it is on earth itself. Therefore sunlight, option 2, is the right answer as sunlight which is the oldest heat source on Earth.

claims:

The model-generated answer makes a logical error in rejecting the possibility of sunlight being the right answer.

Sunlight might be produced in the sun.

Sunlight doesn't stay in the sun.

Sunlight comes to earth.

Sunlight is on earth itself.

Sunlight, option 2, is the right answer as it is the oldest heat source on Earth.

[example 2 end]

[example 3 start]

input:

That the increase has "reached record levels" is not indicated in the context. Otherwise, the answer is adequate, except it should also mention the government is responding to the situation.

claims:

The content that the increase has "reached record levels" is not indicated in the context.

The correct answer should also mention the government is responding to the situation.

[example 3 end]

[example 4 start]

input:

The answer's summary was missing information about Andre Ward winning the rematch.

claims:

The model-generated answer was missing information about Andre Ward winning the rematch

[example 4 end]

~

--------------USER MESSAGE-------------

~

input:

\{critique\}

claims:
\\
\bottomrule
\end{tabular}
    \caption{Prompt for AIUs extraction.}
    \label{tab:prompt_aius_extraction}
\end{table*}

\begin{table*}[t]
    \scriptsize
    \centering
\begin{tabular}{@{}p{\textwidth}@{}}
\toprule
--------------SYSTEM MESSAGE-------------

~

You are a brilliant AI assistant.
You receive an input question, a model-generated answer, the reference answer and a claim that is the critique for the model-generated answer.
To verify the claim, you should think step by step as following:

1) you should consider the input question, the model-generated answer and the reference answer, and then you should determine what you need for verifying the claim. Please note that the input question is the most reliable, and the input question is always correct. The model-generated answer is the least reliable, and it can contain some flaws. The reference answer is more reliable than the model-generated answer, but sometimes it may contain some flaws.

2) verify whether the claim is true or false. If a claim is either a factual statement, explicitly mentioned in the needed information, or logically deducible from it, then the claim is true; otherwise, it is false.

Here are some examples:

[example 1 start]

<input question>

A shelf would fit in what part of a room?
Please choose the correct option.
Here are the options:

Option 1: garage

Option 2: corner

<model-generated answer>

Option 1

<reference answer>

Option 2

<claim>

The model-generated answer is wrong.

<verify claim>

To verify the claim, we need to consider the model-generated answer and the reference answer.
The reference answer says that Option 2 is the correct choice.
The model-generated answer states that the answer is Option 1. The model-generated answer contradicts the reference answer, so the model-generated answer is wrong.
The claim states that the model-generated answer is wrong. The claim is consistent with the above analysis result. Therefore, the claim is true.

[example 1 end]

[example 2 start]

<input question>

Give a summary of the below article:
Biggin Hill Airport wants to increase bank holiday and weekend opening times to increase the number of flights. Director William Curtis said he feared the airport could be asked to help the UK meet increased demand for air travel if the extension was not granted.

<model-generated answer>

Biggin Hill Airport intends to expand its hours to accommodate more flights.

<reference answer>

Biggin Hill Airport seeks to extend opening times to boost flights amidst concerns of meeting UK air travel demand.

<claim>

William Curtis wants to increase opening hours.

<verify claim>

To verify the claim, we need to consider the input question.
The input question says that Biggin Hill Airport wants to increase opening times instead of William Curtis.
The claim states that William Curtis wants to increase opening hours. The claim contradicts with the above reasoning result. Therefore, the claim is false.

[example 2 end]

[example 3 start]

<input question>

Context:
He walked in the door and saw her pointing a gun with a laser pointer at his chest.
Given the above context, How did he feel?

<model-generated answer>

He was glad.

<reference answer>

He was scared for his life.

<claim>

The correct answer is that he was afraid.

<verify claim>

To verify the claim, we need to consider the reference answer.
The reference answer states that he was scared for his life, which is synonymous with being afraid.
The claim states that the correct answer is that he was afraid. The claim is consistent with the above reasoning result. Therefore, the claim is true.

[example 3 end]

[example 4 start]

<input question>

Here is a premise:
A black and white dog sliding down a slope on a beach.

Here is a hypothesis:
A dog going uphill to the beach.

Does this premise imply the hypothesis? Please justify your answer:

<model-generated answer>

No, the dog could have been sliding down any slope, not just a sandy beach.

<reference answer>

No. The premise states that the dog is sliding down a slope on a beach, which indicates a downward movement.
The hypothesis suggests that the dog is going uphill to the beach, which indicates an upward movement.
These two scenarios are contradictory, so the premise does not support or imply the hypothesis.

<claim>

The model-generated answer provides a clear and concise explanation.

<verify claim>

To verify the claim, we need to consider the model-generated answer and the reference answer.
The model-generated answer states that the reason is that the dog could have been sliding down any slope. However, it does not mention the direction of the dog's movement, which is the main point of contradiction.
The claim states that the model-generated answer provides a clear and concise explanation. The claim contradicts with the above reasoning result. Therefore, the claim is false.

[example 4 end]

~

--------------USER MESSAGE-------------

~

<input question>

\{question\}

<model-generated answer>

\{model-generated answer\}

<reference answer>

\{reference answer\}

<claim>

\{AIU from hypothesis critique\}

<verify claim>
\\
\bottomrule
\end{tabular}
    \caption{Prompt for performing the precision task.}
    \label{tab:prompt_precision}
\end{table*}

\begin{table*}[t]
    \scriptsize
    \centering
\begin{tabular}{@{}p{\textwidth}@{}}
\toprule
--------------SYSTEM MESSAGE-------------

~

You are a brilliant AI assistant.
You receive a reference text and a claim. Note that the 'answer' in the reference text means the model-generated answer or the correct answer.
You need to think step by step for verifying whether the claim is true or false. If a claim is either explicitly mentioned in the reference text or logically deducible from it, then the claim is true; otherwise, it is false.

Here are some examples:

[example 1 start]

<reference text>

The model-generated answer gives the wrong step. It is suggested to check the reasoning steps.

<claim>

The model-generated answer is wrong.

<verify claim>

The reference text points that the model-generated answer contains the wrong step, so the model-generated answer is wrong.
The claim states that the model-generated answer is wrong. It is consistent with the above reasoning result. Therefore, the claim is true.

[example 1 end]

[example 2 start]

<reference text>

The model-generated answer is not right, because it provides the wrong analysis. The correct answer should be Option A.

<claim>

The correct answer should point out the invented fact.

<verify claim>

The reference text states that the correct answer should be Option A.
The claim states that the correct answer should point out the invented fact. The claim is not mentioned or implied in the reference text. Therefore, the claim is false.

[example 2 end]

[example 3 start]

<reference text>

The answer lacks detail and context, like the age of the victim, the place of the incident and so on.

<claim>

The accurate summary should contain the age of the dead person.

<verify claim>

The reference text states that the model-generated answer lacks the detail, the age of victim.
The claim states that the correct summary should contain the age of the dead person. It means that the model-generated answer is not mentioned the age of victim. The claim can be logically deducible from the reference text. Therefore, the claim is true.

[example 3 end]

[example 4 start]

<reference text>

The answer could be more concise and focused.

<claim>

The model-generated answer is mostly correct, but it could be improved by providing more specific details.

<verify claim>

The reference text states that the model-generated answer could be more concise. It means that the model-generated answer is elaborated.
The claim states that the model-generated answer could be improved by providing more specific details. It means that the model-generated answer is brief. The claim contracts with the reference text. Therefore, the claim is false.

[example 4 end]

~

--------------USER MESSAGE-------------

~

<reference text>

\{hypothesis critique\}

<claim>

\{AIU from reference critique\}

<verify claim>

\\
\bottomrule
\end{tabular}
    \caption{Prompt for performing the recall task.}
    \label{tab:prompt_recall}
\end{table*}

\begin{table*}[t]
    \scriptsize
    \centering
\begin{tabular}{@{}p{\textwidth}@{}}
\toprule
Your task is to evaluate the generated reference answer (i.e., response for a question). Give a score 1-7 (worst-best) based on the quality of the answer.

7: Perfect. The response perfectly answer the query and provide suitable explanation.

6: Exceptional. The response perfectly answer the query and provide suitable explanation, but introduce some flaws.

5: Excellent. The response correctly answer the query and provide suitable explanation, but introduce some flaws.

4:Good. The response correctly answer the query but lacks suitable explanation.

3: Average. The response incorrectly answer the query but provides consistent explanation.

2: Poor. The response incorrectly answer the query but provides the explanation with flaws.

1: Extremely bad. The generated response is random text or simple repeats the question.

Give a score 1-3 for response with incorrect content and give a score 4-7 for response with correct content.
\\
\bottomrule
\end{tabular}
    \caption{Instruction for human evaluation of generated reference answer.}
    \label{tab:human_eval_answer}
\end{table*}

\begin{table*}[t]
    \scriptsize
    \centering
\begin{tabular}{@{}p{\textwidth}@{}}
\toprule
Your task is to evaluate the critique on a model-generated answer. Give a score 1-7 (worst-best) based on the quality of the critique.

7: When the answer is wrong, the critique clearly highlights the most important errors and provides 	very actionable suggestions. When the answer is correct, the critique confirms the answer is correct and provides very useful suggestions.

6: When the answer is wrong, the critique confirms that the answer is wrong and points out the most important errors. When the answer is correct, the critique confirms the answer is correct and provides useful suggestions.

 5: When the answer is wrong, the critique misses the important errors but clearly confirms that the answer is wrong. When the answer is correct, the critique confirms the answer is correct and proposes some less useful suggestions.

4: The critique has a correct judgment of the answer (e.g., states correct answer is correct or states wrong answer is wrong).

 3: The critique is vague about whether or not the answer is correct. Or the critique itself tries to answer the question regardless of the content in the answer.

2: The critique has a wrong judgment of the answer (e.g., states correct answer is wrong or states wrong answer is correct).

 1: The critique is completely random text or simply repeats the answer.

First, please check whether the critique has correct or incorrect judgment (correct judgment means the answer is correct, critique confirms the correctness. Or if the answer is incorrect, the critique confirms the incorrectness.)

Give a score 1-3 for critique with incorrect judgment and give a score 4-7 for critique with correct judgment.
\\
\bottomrule
\end{tabular}
    \caption{Instruction for human evaluation of generated reference critique.}
    \label{tab:human_eval_critique}
\end{table*}

\begin{table*}[t]
    \scriptsize
    \centering
\begin{tabular}{@{}p{\textwidth}@{}}
\toprule
Your task is to evaluate the AIU extraction on a critique. Give a score 1-7 (worst-best) based on the quality of the extraction.

7: Perfect. Generate all salient claims without extra information.  Useless or redundant information is 	removed.

6: Exceptional. Generate all salient claims without extra information but contains very few useless or 	redundant information.

5: Excellent.  Generate all salient claims but introduce little extra information.

4:Good. Generate some salient claims but introduce little extra information.

3: Average. Generate some salient claims but introduce too much extra information.

2: Poor. Remove too many salient claims, or introduce too much extra information.

1: Extremely bad. The generated claims are random text or simple repeats the critique.
\\
\bottomrule
\end{tabular}
    \caption{Instruction for human evaluation of AIUs extraction.}
    \label{tab:human_eval_aiu}
\end{table*}

\begin{table*}[t]
    \scriptsize
    \centering
\begin{tabular}{@{}p{\textwidth}@{}}
\toprule
--------------SYSTEM MESSAGE-------------

~

You are a brilliant AI assistant. You will receive a question, a model-generated answer, and two critiques about this answer. A good critique should point out key errors contained in the answer and provide constructive suggestions. Your task is to evaluate the quality of the critique. Your evaluation should consider factors such as the accuracy, factuality, comprehensiveness, relevance, and conciseness. Begin your evaluation by comparing the two critiques and provide a short explanation. Avoid any position biases and ensure that the order in which the answers were presented does not influence your decision. Do not allow the length of the critiques to influence your evaluation. Do not favor certain names of the answers. Be as objective as possible. After providing your explanation, output your final verdict by strictly following this format: "[[A]]" if the critique A is better, "[[B]]" if the critique B is better, and "[[C]]" for a tie.

~

--------------USER MESSAGE-------------

~

<input question>

\{question\}

<model-generated answer>

\{model-generated answer\}

<critique A>

\{hypothesis critique from LLM 1\}

<critique B>

\{hypothesis critique from LLM 2\}

\\
\bottomrule
\end{tabular}
    \caption{Prompt for pairwise comparison.}
    \label{tab:pairwise_gpt_prompt}
\end{table*}

\begin{table*}[t]
    \scriptsize
    \centering
\begin{tabular}{@{}p{\textwidth}@{}}
\toprule
--------------SYSTEM MESSAGE-------------

~

You are a brilliant AI assistant. You will receive a question, a model-generated answer, and a critique about this answer. Your task is to evaluate the quality of the critique and give a score.

The score is based on the quality of the critique:

7: When the answer is wrong, the critique clearly highlights the most important errors and provides very actionable suggestions. When the answer is correct, the critique confirms the answer is correct and provides very useful suggestions.

6: When the answer is wrong, the critique confirms that the answer is wrong and points out the most important errors. When the answer is correct, the critique confirms the answer is correct and provides useful suggestions.

5: When the answer is wrong, the critique misses the important errors but clearly confirms that the answer is wrong. When the answer is correct, the critique confirms the answer is correct and proposes some less useful suggestions.

4: The critique has a correct judgement of the answer (e.g., states correct answer is correct or states wrong answer is wrong).

3: The critique is vague about whether or not the answer is correct. Or the critique itself tries to answer the question regardless of the content in the answer.

2: The critique has a wrong judgement of the answer (e.g., states correct answer is wrong or states wrong answer is correct).

1: The critique is completely random text or simply repeats the answer.

Begin your evaluation by considering the critique and provide a short explanation.

First, please check whether the critique has correct or incorrect judgment (correct judgment means the answer is correct, critique confirms the correctness. Or if the answer is incorrect, the critique confirms the incorrectness.) Please note that give a score 1-3 for critique with incorrect judgment and give a score 4-7 for critique with correct judgment.

Be as objective as possible. After providing your explanation, output your final score by strictly following this format: "[[score]]".

~

--------------USER MESSAGE-------------

~

<input question>

\{question\}

<model-generated answer>

\{model-generated answer\}

<critique>

\{hypothesis critique from LLM\}

\\
\bottomrule
\end{tabular}
    \caption{Prompt for single critique rating.}
    \label{tab:single_gpt_prompt}
\end{table*}

\begin{table*}[t]
    \scriptsize
    \centering
\begin{tabular}{@{}p{\textwidth}@{}}
\toprule
--------------SYSTEM MESSAGE-------------

~

You are a brilliant AI assistant.

Based on the provided critique, please refine the model-generated answer to address all issues in the critique and ensure that all suggestions from the critique are incorporated to enhance the refined answer.

Here are some examples:

[example 1 start]

<input question>

A shelf would fit in what part of a room?

Here are the options:

Option 1: garage

Option 2: corner

Please choose the correct option and provide the justification.

<model-generated answer>

Option 1

<critique>

The model-generated answer is wrong and lacks the justification.

<refined answer>

Option 2.

Justification: "Corner" refers to a specific location within a room where two walls intersect, making it the appropriate choice for identifying a part of a room where a shelf could be installed.

[example 1 end]

[example 2 start]

<input question>

Context:

He walked in the door and saw her pointing a gun with a laser pointer at his chest.

Given the above context, How did he feel?

<model-generated answer>

He was pointed by a gun.

<critique>

The model-generated answer is a good starting point, but it should mention his feeling.

<refined answer>

He was scared when he saw a gun.

[example 2 end]

~

--------------USER MESSAGE-------------

~

<input question>

\{question\}

<model-generated answer>

\{model-generated answer\}

<critique>

\{hypothesis critique from LLM\}

<refined answer>
\\
\bottomrule
\end{tabular}
    \caption{Prompt for refining model-generated answer using a critique.}
    \label{tab:refine_gpt_prompt}
\end{table*}

\begin{table*}[t]
    \scriptsize
    \centering
\begin{tabular}{@{}p{\textwidth}@{}}
\toprule
--------------SYSTEM MESSAGE-------------

~

You are a brilliant AI assistant. You can evaluate the quality of the model-generated answer for the input question. Your evaluation should consider factors such as the accuracy, factuality, relevance, and conciseness. Please note that details that do not clearly help to answer the input question are considered redundant. Begin your evaluation by comparing the two answers and provide a short explanation. Avoid any position biases and ensure that the order in which the answers were presented does not influence your decision. Do not allow the length of the answers to influence your evaluation. Do not favor certain names of the answers. Be as objective as possible. After providing your explanation, output your final verdict by strictly following this format: "[[A]]" if the model-generated answer A is better, "[[B]]" if the model-generated answer B is better, and "[[C]]" for a tie.

~

--------------USER MESSAGE-------------

~

<input question>

\{question\}

<model-generated answer A>

\{refined answer from the LLM 1 critique\}

<model-generated answer B>

\{refined answer from the LLM 2 critique\}

\\
\bottomrule
\end{tabular}
    \caption{Prompt for GPT-4 evaluation for a pair of refined answers.}
    \label{tab:eval_answer_gpt}
\end{table*}

\begin{table*}[t]
    \scriptsize
    \centering
\begin{tabular}{@{}p{\textwidth}@{}}
\toprule

Your task is to pick up the better answer from two given answers. Your evaluation should consider factors such as the accuracy, factuality, relevance, and conciseness. Please note that details that do not clearly help to answer the input question are considered redundant. Do not allow the length of the answers to influence your evaluation. Be as objective as possible. Label your final verdict: "A" if the model-generated answer A is better, "B" if the model-generated answer B is better, and "C" for a tie.
\\
\bottomrule
\end{tabular}
    \caption{Instruction for human evaluation for a pair of refined answers.}
    \label{tab:eval_answer_human}
\end{table*}

\end{document}